\documentclass{article}
\usepackage{ijcai17}
\usepackage{times}
\usepackage{graphicx}
\usepackage{amsmath}
\usepackage{amssymb}
\usepackage{url}
\usepackage{caption}
\usepackage{mathrsfs}
\usepackage{algorithm}
\usepackage{algpseudocode}
\usepackage{dsfont}
\usepackage[pagebackref=false,breaklinks=true,letterpaper=true,colorlinks,urlcolor=blue,citecolor=blue,linkcolor=blue,bookmarks=false]{hyperref}
\usepackage{breakurl}
\usepackage{multirow}
\usepackage{booktabs}

\newcommand{\figdir}{figures}
\def\swthree{0.33\columnwidth}

\title{Learning to hallucinate face images via Component Generation and Enhancement}
\author{Yibing Song$^1$, Jiawei Zhang$^1$, Shengfeng He$^2$, Linchao Bao$^3$ and Qingxiong Yang$^4$\\
$^1$City University of Hong Kong\\
$^2$South China University of Technology\\
$^3$Tencent AI Lab\\
$^4$University of Science and Technology of China\\}

\begin{document}

\maketitle

\begin{abstract}
We propose a two-stage method for face hallucination. First, we generate facial components of the input image using CNNs. These components represent the basic facial structures. Second, we synthesize fine-grained facial structures from high resolution training images. The details of these structures are transferred into facial components for enhancement. Therefore, we generate facial components to approximate ground truth global appearance in the first stage and enhance them through recovering details in the second stage. The experiments demonstrate that our method performs favorably against state-of-the-art methods\footnote{Complete experimental results and our implementation are provided on the authors' webpage.}.
\end{abstract}

\section{Introduction}
Face Hallucination (FH) is a domain specific problem which generates high resolution (HR) face images from low resolution (LR) inputs. Different from generic image super resolution (SR) methods, FH exploits specific facial structures and textures. It generates high quality face images compared with generic image SR methods. This activates a series of FH applications ranging from image editing to video surveillance. More generally, FH is taken as a preprocessing step for face related applications.

The state-of-the-art FH methods transfer facial details from HR training images to LR inputs. They aim to exploit the relationship between LR and HR images either globally or locally. One of the solutions is to align face images in pixel-wise precision between the input and training images. So dense correspondences on the training images can be established and HR facial details can be transferred into LR input image in the form of bayesian inference \cite{tappen-eccv12-bayesian} or image gradient \cite{Chih-cvpr13-FH}. The transferred result usually contains more details on the facial component compared with the ones generated using generic image SR techniques.

\renewcommand{\tabcolsep}{.1pt}
\begin{figure}[t]
\begin{center}
\begin{tabular}{ccc}
\vspace{-1mm}\includegraphics[width=0.32\columnwidth]{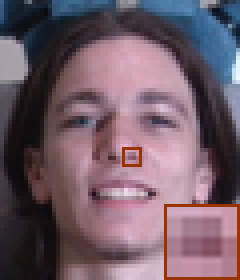}&
\includegraphics[width=0.32\columnwidth]{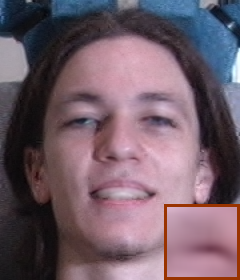}&
\includegraphics[width=0.32\columnwidth]{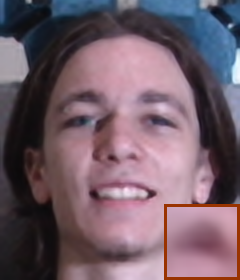}\\
\vspace{-1mm}\footnotesize{(a) Input (NN)}&\footnotesize{(b) SFH}&\footnotesize{(c) SRCNN}\\
\scriptsize{PSNR / SSIM}&\scriptsize{31.13 / 0.85}&\scriptsize{32.98 / 0.89 }\\
\vspace{-1mm}\includegraphics[width=0.32\columnwidth]{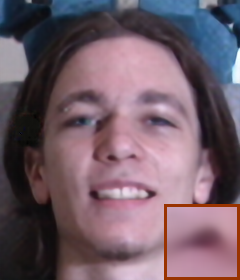}&
\includegraphics[width=0.32\columnwidth]{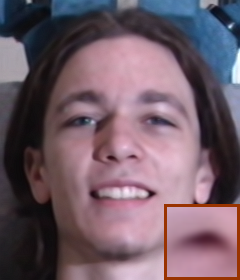}&
\includegraphics[width=0.32\columnwidth]{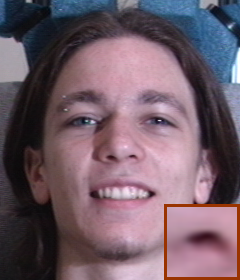}\\
\vspace{-1mm}\footnotesize{(d) SRResNet}&\footnotesize{(e) Ours}&\footnotesize{(f) Ground Truth}\\
\scriptsize{33.47 / 0.89}&\scriptsize{\textbf{34.40} / \textbf{0.90}}&\scriptsize{$+\infty$ / 1}\\
\end{tabular}
\end{center}
\vspace{-5mm}
\caption{The performance of FH and image SR methods.}
\label{fig:intro}
\end{figure}

Despite the demonstrated success, the quality of FH results greatly relies on feature matching between training and input images. Because of the limited texture on the LR input (e.g., $60\times80$), it is difficult to extract handcrafted features such as SIFT \cite{lowe-ijcv04-sift} to make a precise description, especially around facial components (i.e., nose, eyes, and mouth). Such a limitation prevents these features to accurately establish the HR correspondence in the training images. It leads to the incorrect detail transfer and the results will be erroneous. As shown in Fig. \ref{fig:intro}, the nose generated from \cite{Chih-cvpr13-FH} in (b) is in different shape from that of the ground truth in (f).

Recently, Convolutional Neural Network (CNN) has been demonstrated effective in image SR \cite{chao-pami2015-srcnn}. It is formulated as a general form of sparsity representation \cite{jianchao-tip10-scsr} and aims to minimize the pixel-wise difference between network output and ground truth. It achieves state-of-the-art performance on natural images where texture patterns uniformly reside in low frequency base and high frequency details. However, direct applying CNN for FH will blur the facial structure because of the uniqueness of component details. As shown in Fig. \ref{fig:intro}(c) and (d), the results generated using CNN \cite{chao-pami2015-srcnn} or ResNet \cite{ledig-cvpr17-gan} models cannot enrich the high frequency details around noses. Meanwhile, finetuning their model using face images can not make a noticeable improvement. This indicates that CNN based models can not be directly adopted on FH due to the domain specific properties.

In this paper, we Learn to hallucinate face images via Component Generation and Enhancement (\textbf{LCGE}). Different from existing end-to-end CNN networks, we propose a two-stage framework for FH. The first stage learns a mapping function to reconstruct the facial structure of the LR input, which benefits the establishment of HR correspondences. This mapping process is formulated via five CNNs. Each CNN corresponds to one facial component (i.e., eyes, eyebrows, noses, mouth and the remaining region). The input face image is thus divided into five subregions and reconstructed independently using CNN. The advantage of the learned facial component is that the texture information is enriched, which alleviates the matching difficulty of LR images. In the second stage, we generate facial components for both training and input images. And a patch-wise K-NN search is performed for each input component. In this way, we can accurately establish HR correspondences without facial alignment. Then we regress to synthesize HR facial structures with fine grained details. However, the regression is conducted on different subjects, which synthesizes HR structures in different illuminations from our desired output. Finally, the details from the HR structures are transferred to the facial components based on edge-aware image filtering. It can successfully recover the missing details to enhance the components. As a result, the output image well approximates the ground-truth image in both global appearance and facial details.

The contributions of this work are summarized as follows:
\begin{itemize}
  \item We propose to learn deep facial components, which contain basic structure for output and ease the matching difficulty of LR images.
  \item We propose a component enhancement method. The fine grained facial structures can be effectively extracted from training dataset and their details will be transferred to enhance deep components.
  \item Quantitative evaluations on the standard benchmarks indicate that the proposed method performs favorably against state-of-the-art approaches.
\end{itemize}

\begin{figure*}[t]
\begin{center}
\begin{tabular}{c}
\includegraphics[width=0.7\linewidth]{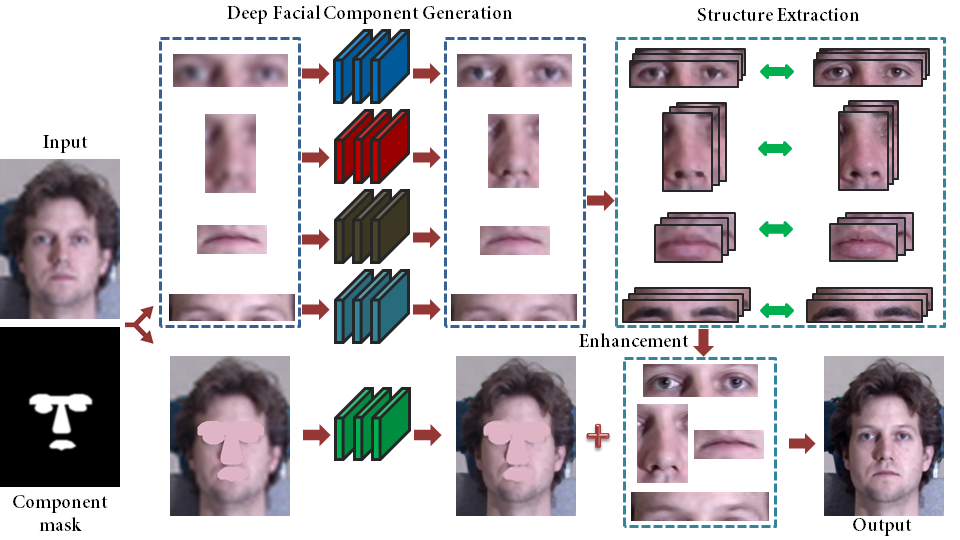}\\
\end{tabular}
\end{center}
\vspace{-7mm}
\caption{Pipeline of the LCGE algorithm. The LR input image is divided into five facial components. Each of them is upsampled using corresponding CNN to generate deep facial component. Fine grained structures can be extracted from HR training images. We transfer their details to enhance deep facial component to generate output result.}
\label{fig:pipeline}
\end{figure*}

\section{Related Work}

Learning based framework is widely adopted in FH methods \cite{wang-ijcv14-survey,song_eccv14_sketch,wang-tip17-bayesian}. They aim to learn the transformation between LR and HR to recover the missing details from the input. In \cite{gunturk-tip03-eigenface,wang-SMC05-hf} generalized approaches on eigen domain are proposed to map both LR and HR image spaces. Tensor based approaches are introduced in \cite{liu-cvpr05-hf,jia-tip08-generalized}. They can well upsample multiple model face images across different poses and expressions. In \cite{liu-ijcv07-FH} Principle Component Analysis (PCA) based linear constraints are learned from training images and a patch-based Markov Random Field (MRF) is used to reconstruct the residues. Instead of directly using patch match \cite{xiang-pr10-FH} to find correspondence, FH methods adopt image alignment where HR images are matched to LR ones by SIFT flow \cite{tappen-eccv12-bayesian} or gradient \cite{Chih-cvpr13-FH}. The quality of output results depends on image alignment, which sometimes fails when poses and expressions are different between training and input images. The convolution neural networks have been adopted in image SR \cite{chao-pami2015-srcnn,kim-cvpr16-deeply} and FH \cite{zhou-aaai15-learning,yu-eccv16-ultra}. Different from existing methods, ours takes the superior performance of CNN to model global appearance and enriches local details through feature matching. It combines the advantage of image SR and FH methods to improve the face image quality.

\section{Proposed Algorithm}

We present the pipeline of LCGE in Fig. \ref{fig:pipeline}. We use CNN to generate deep facial components for the input LR image. They contain basic structure of the output while details are not recovered completely. These components benefit the establishment of LR-HR correspondences and thus fine grained structures can be effectively extracted. The details of these structures are added back to enhance deep facial components to generate the output result.

\subsection{Deep Facial Component Generation}\label{sec:cnn}

We categorize face image into five subregions. Four of them are defined as facial components covering eyes, eyebrows, noses and mouths. The last one is defined as the remaining region. These subregions can be easily obtained using component mask generated by facial landmarks. For an input LR image, we first upsample it to the same resolution as the output using bicubic interpolation and obtain five subregion patches. Then we take each patch as input to the corresponding CNN to generate deep facial component. We have five CNNs in total, each of them contains three convolutional layers. The network structure and training process are similar with those of SRCNN \cite{chao-pami2015-srcnn}.

\subsubsection{Discussion}
We generate the deep facial components for two purposes. First, CNN is effective to minimize the pixel difference between its output and the ground truth. We divide face image into different components and train one CNN for each component independently. Each CNN is set to capture the specific feature of one facial component and generate basic structures of output. Meanwhile, deep facial component is set as an intermediate state between bicubic upsampling of LR input and the ground truth HR image. It is effective to recover the majority of basic structures except some tiny high frequency details. So the remaining work aims to capture such missing details to enhance deep facial component. In this way, the output will approximate ground truth in both global appearance and local details.

Second, deep facial components are able to transform both input and training images into a similar condition, which enables the accurate establishment of HR correspondences so that fine grained facial structures can be effectively extracted. We downsample facial components from HR training images as input. So we can generate deep facial counterpart for each facial component of training images and formulate a training pair with the HR corresponding component. The training pairs formulation is effective to synthesize HR facial structure through component searching. We compare the similarity of the deep facial component between LR input and LR training images. Once similar components are identified we locate the corresponding HR facial components. Different from the prior art which performs feature matching between LR of input and training images, deep facial component enriches facial texture information and thus can accurately establish the HR correspondences. We use intensity and structure based metric for matching (as shown in Eq. \ref{eq:ncc_abs}) and find it performs well in practice. The main reason is that deep facial component is descriptive enough to distinguish the ambiguity from LR. As such, there is no need to use SIFT \cite{lowe-ijcv04-sift} or CNN features \cite{rbg-cvpr14-RCNN}.

\subsection{Component Enhancement}\label{sec::structure_enhancement}
Although deep facial component generation enriches structure information for LR input patches, blur effect still occurs and high frequency details cannot be recovered. Here we propose a component enhancement method to recover high frequency details for the components. It consists of two steps. First, we extract fine grained facial structure from preconstructed training pairs. Then we transfer structure details to enhance deep facial component to generate the output.

\begin{figure}[]
\begin{center}
\begin{tabular}{c}
\includegraphics[width=\linewidth]{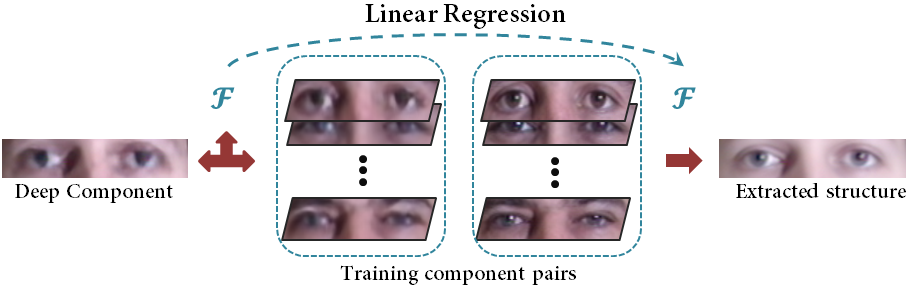}\\
\end{tabular}
\end{center}
\vspace{-5mm}
\caption{Structure extraction through dataset synthesis. We use CNN to generate deep facial component for both training and input images. Then for each input component patch we establish correspondences from the training dataset. We learn a linear regression function $F$ through the component patches, and use $F$ to map HR training patches to generate extracted result.}
\label{fig:structure_extraction}
\end{figure}

\renewcommand{\tabcolsep}{.1pt}
\begin{figure*}[t]
\def\wfour{0.24\linewidth}
\begin{center}
\begin{tabular}{cccc}
\vspace{-1mm}\includegraphics[width=\wfour]{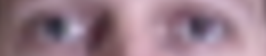}&
\includegraphics[width=\wfour]{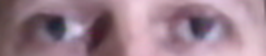}&
\includegraphics[width=\wfour]{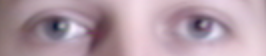}&
\includegraphics[width=\wfour]{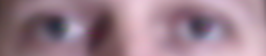}\\
\small{(a) Input}&\small{(b) Deep facial component}&\small{(c) Extracted structure}&\small{(d) GF on (b) using (c)}\\
\vspace{-1mm}\includegraphics[width=\wfour]{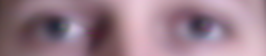}&
\includegraphics[width=\wfour]{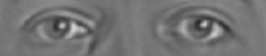}&
\includegraphics[width=\wfour]{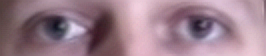}&
\includegraphics[width=\wfour]{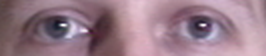}\\
\small{(e) GF on (c) using (c)}&\small{(f) Detail: (c)-(e)}&\small{(g) Output: (d)+(f)}&\small{(h) Ground Truth}\\
\end{tabular}
\end{center}
\vspace{-5mm}
\caption{Texture transfer. Input LR face image is shown in (a). We generate deep structure shown in (b). Extracted texture is shown in (c). We perform guided filtering on (b) using (c) as guidance shown in (d). We also filter (c) using guided filtering in (e). The lost facial details through filtering can be identified in (f), which is the difference between (c) and (e). We add the details back to (c) to generate the output shown in (g). The ground truth image is shown in (h).}
\label{fig:structure_transfer}
\end{figure*}

\subsubsection{Structure Extraction}
We aim to extract facial structure from HR training images where the subjects are different from that on an input image. Inspired by \cite{hertzmann-siggraph01-analogy} which involves training image pairs to transfer image style, we construct a training component dataset for facial structure extraction. For each categorized component of the training images, we downsample it into LR and upsample using bicubic interpolation. Then we use the upsampled component as input to obtain deep facial component. As a result training component pairs can be generated which consist of well aligned deep facial components and corresponding HR components.

Given an input image we divide it into different components represented by local patches. For one patch centered on pixel $p$, we perform a K nearest neighbor search (K-NN) on the deep facial component of the training pairs to find the corresponding patches. The patch similarity metric is defined as the combination of normalized cross correlation $D_{ncc}$ and absolute difference $D_{abs}$:
\begin{equation}
D_p=\alpha\cdot(1-D_{ncc})+(1-\alpha)\cdot D_{abs},
\label{eq:ncc_abs}
\end{equation}
where $\alpha$ is set as 0.2 and K is set as 5 in our experiments. We normalize image pixel value to $[0,1]$ in order to set two metrics into the same range.

After K-NN search we select $K$ candidate patches from deep components. Let $\bar{\mathrm{T}}_p^i$ ($i\in[1,\cdots,K]$) denote one vector containing all the pixel values of the $i$th candidate patch, and $\bar{\mathrm{I}}_p$ denote a vector containing the pixel values of the input patch. We also denote the linear regression function as $\mathcal{F}_p=[F_p^1,\cdots,F_p^K]^\mathrm{T}$ where $F_p^i$ ($i\in[1,\cdots,K]$) is each coefficient of $\mathcal{F}_p$. The energy function is defined as:
\begin{equation}
E_p^{\mathrm{data}}=||\bar{\mathrm{T}}_p\cdot\mathcal{F}_p-\bar{\mathrm{I}}_p||^2,
\label{eq:energy_data}
\end{equation}
where $\bar{\mathrm{T}}_p=[\bar{\mathrm{T}}_p^1,\bar{\mathrm{T}}_p^2,\cdots,\bar{\mathrm{T}}_p^K]$. It is a linear regression problem and we can compute $\mathcal{F}_p$ as
\begin{equation}
\mathcal{F}_p=(\bar{\mathrm{T}}_p^\mathrm{T}\cdot\bar{\mathrm{T}}_p)^{-1}\bar{\mathrm{T}}_p\cdot\bar{\mathrm{I}}_p.
\label{eq:solver}
\end{equation}
We can efficiently compute $\mathcal{F}_p$ when the patches contain texture (i.e., the pixel values in $\bar{\mathrm{T}}_p$ should not be similar with each other). However, in some cases when $p$ is on the smooth region (e.g, nose) $\bar{\mathrm{T}}_p^\mathrm{T}$ may be singular and thus $\mathcal{F}_p$ becomes outliers. We resolve the problem by adding a regularization term as:
\begin{equation}
E_p = E_p^{\mathrm{data}}+E_p^{reg} = ||\bar{\mathrm{T}}_p\cdot\mathcal{F}_p-\bar{\mathrm{I}}_p||^2+\lambda ||\mathcal{F}_p||^2,
\end{equation}
where $\lambda$ is the weight controlling the influence of regularization term. It is set as the number of pixels in input patch. We can solve the above energy function as:
\begin{equation}
\mathcal{F}_p=(\bar{\mathrm{T}}_p^\mathrm{T}\cdot\bar{\mathrm{T}}_p+\lambda\mathds{1})^{-1}\bar{\mathrm{T}}_p\cdot\bar{\mathrm{I}}_p,
\label{eq:solver_final}
\end{equation}
where $\mathds{1}$ is the identity matrix.

Once we calculate the regression function $\mathcal{F}_p$, we map the HR training patches into the extracted patch. Let $\textrm{T}_p^i$ ($i\in[1,\cdots,K]$) denote one vector containing the pixel values of the corresponding HR training patches. The extracted patch $\textrm{R}_p$ can be computed as:
\begin{equation}
\textrm{R}_p=\sum_{i=1}^K{F_p^i\cdot\textrm{T}_p^i}.
\label{eq:mapping}
\end{equation}
We compute the extracted patch for each pixel on the input patch. For the overlapping areas between different patches, we perform averaging to generate the result shown in Fig. \ref{fig:structure_extraction}. It can effectively extract fine grained structures through synthesizing from the HR training images.

\subsubsection{Detail Transfer}
The extracted facial structure contains high frequency details lost in the deep facial component. However, it can not be directly adopted as the output. This is because we extract structure from several training patches which belong to different subjects. The illumination of each subject is different from each other, which results in different grayscale values between extracted structure and ground truth (e.g., Fig. \ref{fig:structure_transfer} (c) and (h)). We notice that the missing details mostly reside in high frequency (e.g., eyes in Fig. \ref{fig:structure_transfer}). To recover the missing details to enhance deep facial component, we propose a detail transfer method based on edge-preserving filtering \cite{Petschnigg-siggraph04-JBF,Eisemann-siggraph04-JBF}. It can effectively extract the missing details and transfer them back to the deep facial component.

The main steps of detail transfer are shown in Fig. \ref{fig:structure_transfer}. We have a deep facial component patch shown in (b) and a extracted structure shown in (c). We use guided filter \cite{kaiming-pami2013-GuidedFilter} to smooth (b) using (c) as guidance. As such, the facial structure of (c) can be transferred into (b). However, the filtered result is likely to be smoothed (as shown in Fig. \ref{fig:structure_transfer} (d)) through guided filtering process. Nevertheless, we can capture the missing details with the help of (c) to create a similar blurry scenario. First, we smooth (c) using guided filtering with itself as guidance shown in (e). Then missing facial details can be captured through subtracting the smoothed image using (c). As shown in (f), the missing details mainly reside around facial components (e.g, eyes). We add (f) to (d) to recover the missing facial details shown in (g). As a result, both global appearance and facial details of the output component patch is similar to the ground truth shown in (h). After we transfer all the component patches we combine them to generate the output face image.

\section{Experiments}

We conduct experiments on four datasets: Multi-PIE \cite{gross-ivc10-multiPie} frontal, Multi-PIE pose, PubFig \cite{kumar-iccv09-pubFig} and Multi-PIE HR datasets. In the Multi-PIE pose dataset, face images are taken with pose around 45 degrees while in the other datasets all face images are taken in frontal view. In the PubFig datasets, input images are captured in real world wild condition while in other datasets the inputs are in the lab controlled environment. The resolution of ground truth images in all datasets except Multi-PIE HR is 320$\times$240, and we set the scaling factor as 4. In Multi-PIE HR dataset the resolution of HR images is 800$\times$600, and we set the scaling factor as 10 to evaluate the performance of different algorithms in such an extreme case.

\def\pp{\hspace{0mm}}
\renewcommand{\tabcolsep}{5pt}
\begin{table}[t]
\caption{Multi-PIE Frontal Dataset}
\centering
       \vspace{-3mm}\begin{tabular}{cccccccc}
        \toprule
        \scriptsize{}&\scriptsize{Bicubic}&\scriptsize{FHTP}\pp&
        \pp\scriptsize{SRSC}&\pp\scriptsize{SFH}&\pp\scriptsize{SRCNN}
        &\pp\scriptsize{SRResNet}&\pp\scriptsize{Ours}\\
        \midrule
        \scriptsize{PSNR}&\scriptsize{32.43}
        &\scriptsize{30.13}\pp&\pp\scriptsize{33.54}&\pp\scriptsize{31.60}&\pp\scriptsize{33.89}&\pp\scriptsize{34.10}&\pp\scriptsize{\textbf{35.17}}\\
        \scriptsize{SSIM}&\scriptsize{0.89}
        &\scriptsize{0.82}\pp&\scriptsize{0.90}&\pp\scriptsize{0.86}&\pp\scriptsize{0.90}&\pp\scriptsize{0.90}&\pp\scriptsize{\textbf{0.92}}\\
        \bottomrule
       \end{tabular}
\label{tab:frontal}
\caption{Multi-PIE Pose Dataset}
\centering
       \vspace{-3mm}\begin{tabular}{cccccccc}
        \toprule
        \scriptsize{}&\scriptsize{Bicubic}&\scriptsize{FHTP}\pp
        &\pp\scriptsize{SCSR}&\pp\scriptsize{SFH}&\pp\scriptsize{SRCNN}
        &\pp\scriptsize{SRResNet}&\pp\scriptsize{Ours}\\
        \midrule
        \scriptsize{PSNR}&\scriptsize{33.97}&\scriptsize{24.59}\pp&\pp\scriptsize{35.14}&\pp\scriptsize{32.84}&\pp\scriptsize{35.45}&\pp\scriptsize{35.64}&\pp\scriptsize{\textbf{37.55}}\\
        \scriptsize{SSIM}&\scriptsize{0.90}&\scriptsize{0.72}\pp&\pp\scriptsize{0.91}&\pp\scriptsize{0.86}&\pp\scriptsize{0.91}&\pp\scriptsize{0.92}&\pp\scriptsize{\textbf{0.94}}\\
        \bottomrule
       \end{tabular}
\label{tab:pose}
\caption{PubFig Dataset}
\centering
       \vspace{-3mm}\begin{tabular}{cccccccc}
        \toprule
        \scriptsize{}&\scriptsize{Bicubic}&\scriptsize{FHTP}\pp
        &\pp\scriptsize{SCSR}&\pp\scriptsize{SFH}&\pp\scriptsize{SRCNN}
        &\pp\scriptsize{SRResNet}&\pp\scriptsize{Ours}\\
        \midrule
        \scriptsize{PSNR}&\scriptsize{29.55}\pp&\pp\scriptsize{26.56}&\pp\scriptsize{30.74}&\pp\scriptsize{28.51}&\pp\scriptsize{31.03}&\pp\scriptsize{31.23}&\pp\scriptsize{\textbf{31.70}}\\
        \scriptsize{SSIM}&\scriptsize{0.86}\pp&\pp\scriptsize{0.71}&\pp\scriptsize{0.88}&\pp\scriptsize{0.82}&\pp\scriptsize{0.88}&\pp\scriptsize{0.88}&\pp\scriptsize{\textbf{0.89}}\\
        \bottomrule
       \end{tabular}
\label{tab:PubFig}
\caption{Multi-PIE HR Dataset}
\centering
       \vspace{-3mm}\begin{tabular}{cccccccc}
        \toprule
        \scriptsize{}&\scriptsize{Bicubic}&\scriptsize{FHTP}\pp
        &\pp\scriptsize{SCSR}&\pp\scriptsize{SFH}&\pp\scriptsize{SRCNN}
        &\pp\scriptsize{SRResNet}&\pp\scriptsize{Ours}\\
        \midrule
        \scriptsize{PSNR}&\scriptsize{30.38}&\scriptsize{24.98}\pp&\pp\scriptsize{30.50}&\pp\scriptsize{28.71}&\pp\scriptsize{30.41}&\pp\scriptsize{30.72}&\pp\scriptsize{\textbf{31.24}}\\
        \scriptsize{SSIM}&\scriptsize{0.73}&\scriptsize{0.66}\pp&\pp\scriptsize{0.75}&\pp\scriptsize{0.64}&\pp\scriptsize{0.75}&\pp\scriptsize{0.74}&\pp\scriptsize{\textbf{0.76}}\\
        \bottomrule
       \end{tabular}
\label{tab:hr}
\caption*{\small{Quantitative evaluations on benchmark datasets. Our method performs favorably against state-of-the-art methods in general.}}
\end{table}

In Multi-PIE frontal dataset, we keep the same setting with that in \cite{Chih-cvpr13-FH} where 2184 images are taken as training and 342 images are taken as input. For Multi-PIE pose and Multi-PIE HR datasets, we adopt leave-one-out strategy for 84 images and 249 images, respectively. For pubFig dataset, we use training images from Multi-PIE frontal to generate 400 output images, which indicates the generality of each method for the real world images. The proposed LCGE method is compared with the state-of-the-art FH methods including FHTP~\cite{liu-ijcv07-FH}, SFH~\cite{Chih-cvpr13-FH} and four image SR methods including bicubic interpolation, SCSR~\cite{jianchao-tip10-scsr}, SRCNN~\cite{chao-pami2015-srcnn} and SRResNet~\cite{ledig-cvpr17-gan}. PSNR and SSIM \cite{wang-tip04-SSIM} are used to measure image quality.

\renewcommand{\tabcolsep}{.1pt}
\def\swthree{0.32\linewidth}
\begin{figure}[t]
\begin{center}
\begin{tabular}{ccc}
    \vspace{-1mm}\includegraphics[width=\swthree]{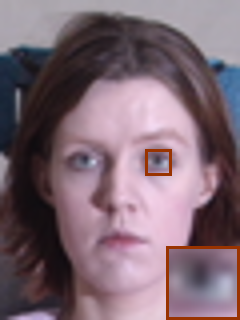}&
    \includegraphics[width=\swthree]{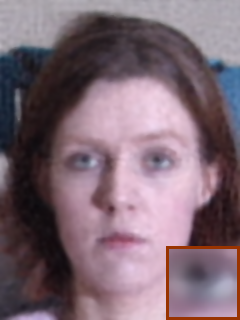}&
    \includegraphics[width=\swthree]{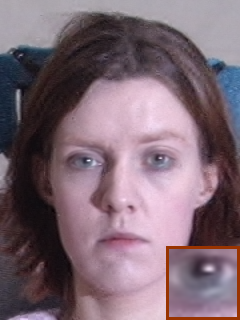}\\
    \vspace{-1mm}\small{(a) Input (Bic)}&\small{(b) FHTP}&\small{(c) SFH}\\
    \scriptsize{32.29 / 0.88}&\scriptsize{29.96 / 0.81}&\scriptsize{31.41 / 0.85}\\
    \vspace{-1mm}\includegraphics[width=\swthree]{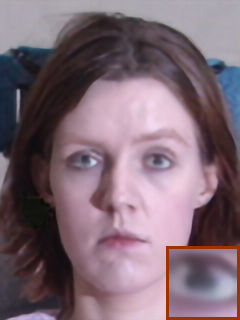}&
    \includegraphics[width=\swthree]{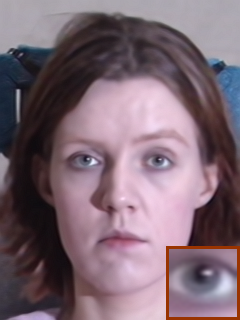}&
    \includegraphics[width=\swthree]{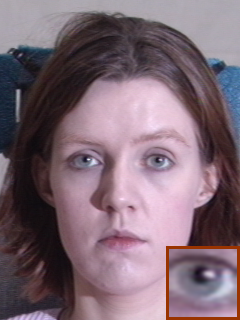}\\
    \vspace{-1mm}\small{(d) SRResNet}&\small{(e) Ours}&\small{(f) Ground Truth}\\
    \scriptsize{34.45 / 0.90}&\scriptsize{\textbf{35.34} / \textbf{0.91}}&\scriptsize{PSNR / SSIM}\\
\end{tabular}
\end{center}
\vspace{-5mm}
\caption{Qualitative evaluation for 4$\times$ upsampled face images in Multi-PIE frontal dataset.}
\label{fig:multiPie1}
\end{figure}

Table \ref{tab:frontal} reports the quantitative performance on Multi-PIE frontal dataset under each metric. It shows that bicubic interpolation achieves higher PSNR value than existing FH methods (i.e, FHTP and SFH). This is because FH methods establish HR correspondences through image alignment which is based on hand crafted features such as SIFT flow \cite{liu-pami11-siftflow}. As the resolution of the input image is low, existing handcrafted features cannot accurately locate HR correspondences. So mismatch occurs and incorrect facial structure will be transferred. As a result, around facial component areas, we will find the distortion of the shape, shifting of the location or change of the lightness, as shown in Fig. \ref{fig:multiPie1} (b) and (c). These artifacts deteriorate the image quality. The SCSR, SRCNN and SRResNet methods achieve high PSNR values due to their global optimization scheme. However, blur occurs around high frequency facial components including eyes, noses, and mouth, which limits the image quality as well. The proposed LCGE method recovers the original image content in both low and high frequencies. It enables the similarity of global appearance and local details, which leads to higher numerical values. The remaining datasets indicate similar quantitative performance in Table \ref{tab:pose}-Table \ref{tab:hr}. SCSR, SRCNN, and SRResNet are shown to favor better numerical scores than FH methods. But they are still not as good as the performance of proposed LCGE method.

The qualitative evaluation is shown in Fig. \ref{fig:multiPie1}. The result of FHTP shown in (b) contains noisy and ghosting artifacts (e.g, facial skin) as well as over smoothed facial components (e.g, eyes). The image SR method SRResNet can achieve high numerical scores because of the global optimization scheme. However, they cannot capture high frequency facial details. As shown in (d), the eyeball and eyelid are blurred, as well as noses and mouths. In comparison, SFH can generate high quality facial components shown in (c). This is because SFH selects the most similar component from the dataset and transfer its gradient to recover high frequency details. However, the facial component correspondence can not be well established in LR. In this case, gradient transfer leads to the dissimilar generation of the facial component. The lighting, shape, and position of the left eye in (c) is different from that in (f) in the close ups although they look similar. In addition, noise is included due to incorrect matching around the mouth region. This limitation is solved by the proposed LCGE method where we synthesize from HR images. Through regression we can correctly generate fine grained structures and transfer their details back to the deep facial component. As a result, LCGE will maintain facial details and thus achieve better quantitative values shown in (g). In addition, Fig. \ref{fig:multiPie2} and \ref{fig:pubfig} demonstrate similar performance in varying pose and real world conditions, respectively.

\renewcommand{\tabcolsep}{.1pt}
\def\swthree{0.32\linewidth}
\begin{figure}[!ht]
\begin{center}
\begin{tabular}{ccc}
    \vspace{-1mm}\includegraphics[width=\swthree]{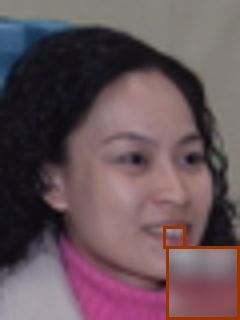}&
    \includegraphics[width=\swthree]{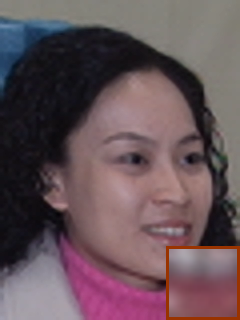}&
    \includegraphics[width=\swthree]{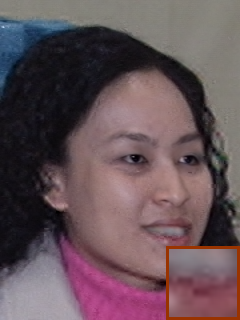}\\
    \vspace{-1mm}\small{(a) Input (Bic)}&\small{(b) SCSR}&\small{(c) SFH}\\
    \scriptsize{34.54 / 0.92}&\scriptsize{35.61 / 0.93}&\scriptsize{33.20 / 0.88}\\
    \vspace{-1mm}\includegraphics[width=\swthree]{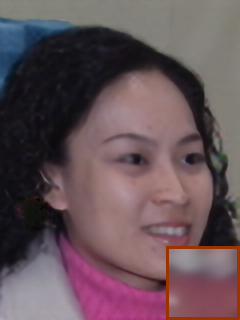}&
    \includegraphics[width=\swthree]{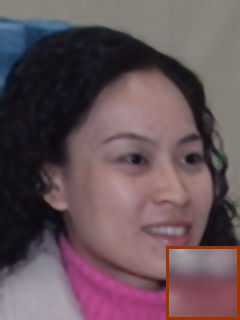}&
    \includegraphics[width=\swthree]{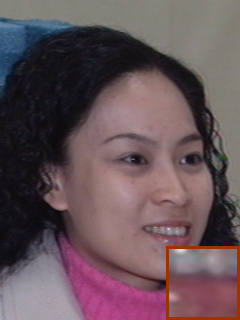}\\
    \vspace{-1mm}\small{(d) SRResNet}&\small{(e) Ours}&\small{(f) Ground Truth}\\
    \scriptsize{36.38 / 0.94}&\scriptsize{\textbf{38.04} / \textbf{0.95}}&\scriptsize{PSNR / SSIM}\\
\end{tabular}
\end{center}
\vspace{-5mm}
\caption{Qualitative evaluation for 4$\times$ upsampled face images in Multi-PIE pose dataset.}
\label{fig:multiPie2}
\begin{center}
\begin{tabular}{ccc}
    \vspace{-1mm}\includegraphics[width=\swthree]{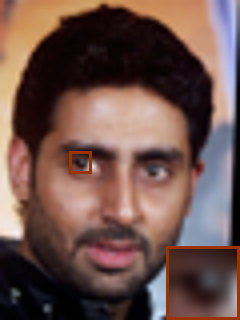}&
    \includegraphics[width=\swthree]{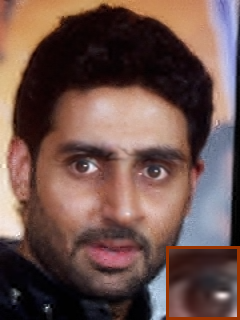}&
    \includegraphics[width=\swthree]{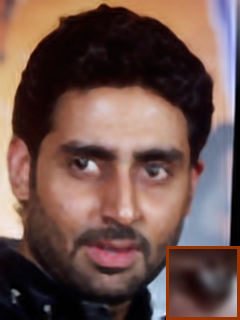}\\
    \vspace{-1mm}\small{(a) Input (Bic)}&\small{(b) SFH}&\small{(c) SRCNN}\\
    \scriptsize{29.41 / 0.91}&\scriptsize{28.82 / 0.87}&\scriptsize{31.25 / 0.93}\\
    \vspace{-1mm}\includegraphics[width=\swthree]{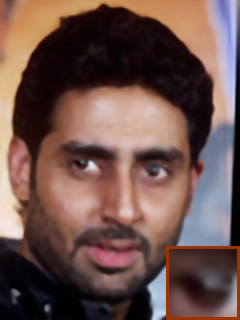}&
    \includegraphics[width=\swthree]{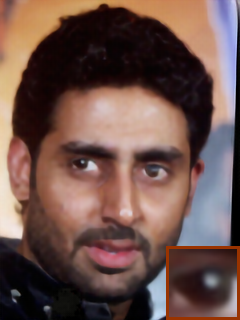}&
    \includegraphics[width=\swthree]{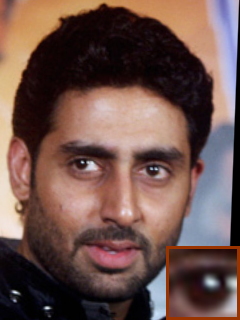}\\
    \vspace{-1mm}\small{(d) SRResNet}&\small{(e) Ours}&\small{(f) Ground Truth}\\
    \scriptsize{32.11 / 0.93}&\scriptsize{\textbf{33.01} / \textbf{0.94}}&\scriptsize{PSNR / SSIM}\\
\end{tabular}
\end{center}
\vspace{-5mm}
\caption{Qualitative evaluation for 4$\times$ upsampled face images in PubFig dataset.}
\label{fig:pubfig}
\end{figure}

\renewcommand{\tabcolsep}{.1pt}
\def\swtwo{0.48\linewidth}
\begin{figure}[!t]
\begin{center}
\begin{tabular}{cc}
\vspace{-1mm}\includegraphics[width=\swtwo]{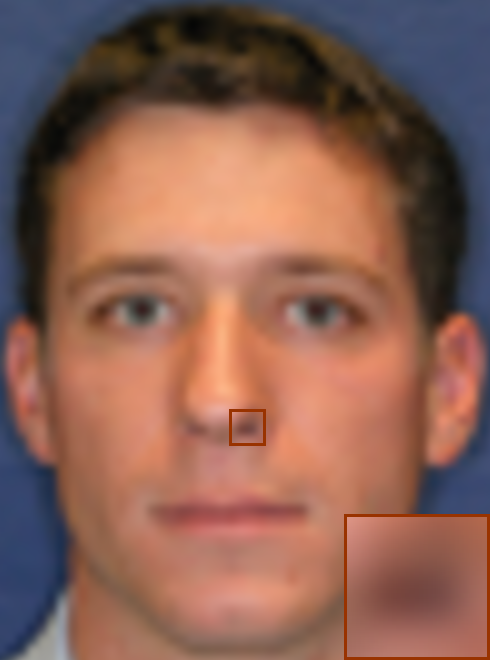}&
\includegraphics[width=\swtwo]{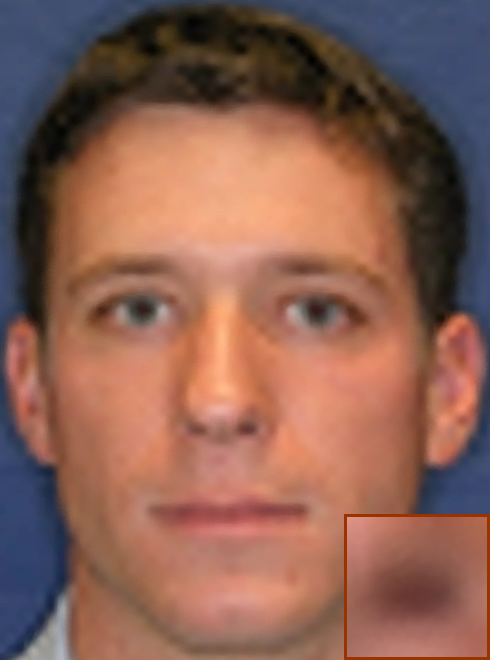}\\
\vspace{-1mm}\small{(a) Input (Bic)}&\small{(b) SCSR}\\
\scriptsize{30.76 / 0.74}&\scriptsize{31.66 / 0.78}\\
\vspace{-1mm}\includegraphics[width=\swtwo]{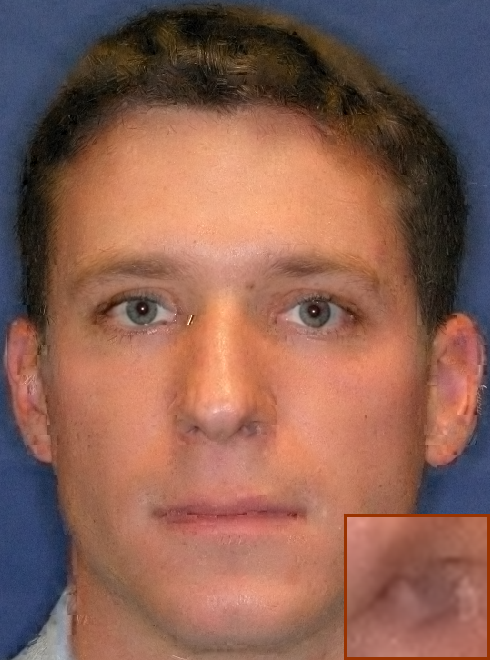}&
\includegraphics[width=\swtwo]{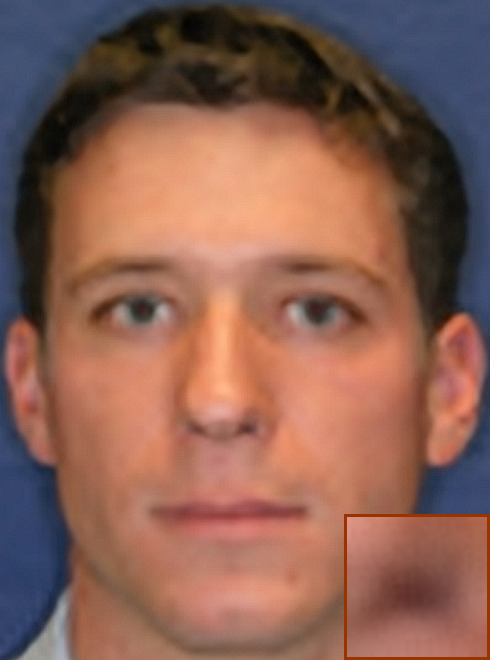}\\
\vspace{-1mm}\small{(c) SFH}&\small{(d) SRCNN}\\
\scriptsize{29.60 / 0.67}&\scriptsize{31.69 / 0.77}\\
\vspace{-1mm}\includegraphics[width=\swtwo]{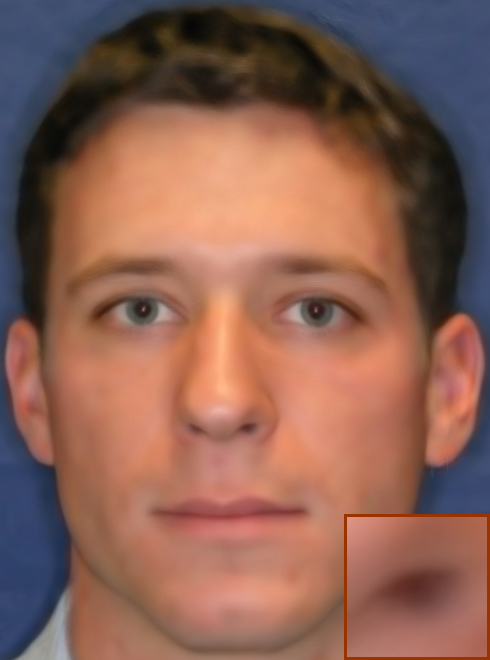}&
\includegraphics[width=\swtwo]{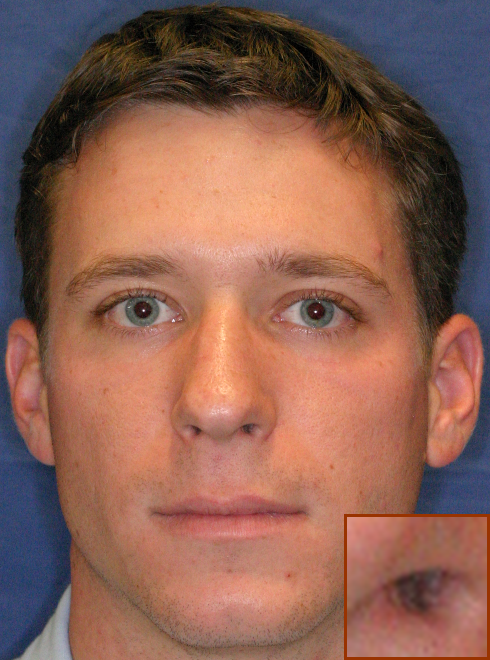}\\
\vspace{-1mm}\small{(e) Ours}&\small{(f) Ground Truth}\\
\scriptsize{\textbf{32.43} / \textbf{0.79}}&\scriptsize{PSNR / SSIM}\\
\end{tabular}
\end{center}
\vspace{-5mm}
\caption{Qualitative evaluation for 10$\times$ upsampled face image in Multi-PIE HR dataset.}
\label{fig:HR}
\end{figure}

The proposed LCGE performs favorably against existing methods in large scaling factors. As shown in Fig. \ref{fig:HR} the evaluation is conducted under upscaling factor of 10, which is not conducted by previous FH methods. The visual performance indicates SCSR and SRCNN produce blur on the results shown in (b) and (d). It is because under such a high upscaling factor sparse coding and CNN based methods can not model the relationship between LR and HR well. The result obtained from SFH in (c) contains high frequency details (e.g, eye) when facial components are correctly matched. However, artifacts occur on the mismatched components (e.g., nose and mouth). In comparison, LCGE generates high quality facial structures through HR synthesis and transferring their details to enhance deep component, which maintains high quality global appearance and facial details shown in (e).

\section{Concluding Remarks}
We propose a FH method named LCGE which integrates global appearance modeling and local feature matching. Different from existing FH methods which adopt handcrafted features for patch matching, LCGE generates deep facial components to narrow down the gap between LR input and HR correspondences. As such, the facial texture is enriched, which eases the matching difficulty. Then fine grained facial structure can be effectively extracted and their details are transferred back to generate the output result. Extensive experiments demonstrate the effectiveness of the proposed LCGE method compared with state-of-the-art approaches.

\clearpage
\small
\bibliographystyle{named}
\bibliography{ref}

\begin{thebibliography}{}

\bibitem[\protect\citeauthoryear{Dong \bgroup \em et al.\egroup
  }{2015}]{chao-pami2015-srcnn}
Chao Dong, Chen~Change Loy, Kaiming He, and Xiaoou Tang.
\newblock Image super-resolution using deep convolutional networks.
\newblock {\em IEEE Transactions on Pattern Analysis and Machine Intelligence},
  2015.

\bibitem[\protect\citeauthoryear{Eisemann and
  Durand}{2004}]{Eisemann-siggraph04-JBF}
Elmar Eisemann and Fr{\'e}do Durand.
\newblock Flash photography enhancement via intrinsic relighting.
\newblock {\em ACM Transactions on Graphics (SIGGRAPH)}, 2004.

\bibitem[\protect\citeauthoryear{Girshick \bgroup \em et al.\egroup
  }{2014}]{rbg-cvpr14-RCNN}
Ross Girshick, Jeff Donahue, Trevor Darrell, and Jitendra Malik.
\newblock Rich feature hierarchies for accurate object detection and semantic
  segmentation.
\newblock In {\em IEEE conference on computer vision and pattern recognition},
  2014.

\bibitem[\protect\citeauthoryear{Gross \bgroup \em et al.\egroup
  }{2010}]{gross-ivc10-multiPie}
Ralph Gross, Iain Matthews, Jeffrey Cohn, Takeo Kanade, and Simon Baker.
\newblock Multi-pie.
\newblock {\em Image and Vision Computing}, 2010.

\bibitem[\protect\citeauthoryear{Gunturk \bgroup \em et al.\egroup
  }{2003}]{gunturk-tip03-eigenface}
Bahadir~K Gunturk, Aziz~U Batur, Yucel Altunbasak, Monson~H Hayes, and
  Russell~M Mersereau.
\newblock Eigenface-domain super-resolution for face recognition.
\newblock {\em IEEE Transactions on Image Processing}, 2003.

\bibitem[\protect\citeauthoryear{He \bgroup \em et al.\egroup
  }{2013}]{kaiming-pami2013-GuidedFilter}
Kaiming He, Jian Sun, and Xiaoou Tang.
\newblock Guided image filtering.
\newblock {\em IEEE Transactions on Pattern Analysis and Machine Intelligence},
  2013.

\bibitem[\protect\citeauthoryear{Hertzmann \bgroup \em et al.\egroup
  }{2001}]{hertzmann-siggraph01-analogy}
Aaron Hertzmann, Charles~E Jacobs, Nuria Oliver, Brian Curless, and David~H
  Salesin.
\newblock Image analogies.
\newblock {\em ACM Transactions on Graphics (SIGGRAPH)}, 2001.

\bibitem[\protect\citeauthoryear{Jia and Gong}{2008}]{jia-tip08-generalized}
Kui Jia and Shaogang Gong.
\newblock Generalized face super-resolution.
\newblock {\em IEEE Transactions on Image Processing}, 2008.

\bibitem[\protect\citeauthoryear{Kim \bgroup \em et al.\egroup
  }{2016}]{kim-cvpr16-deeply}
Jiwon Kim, Jung Kwon~Lee, and Kyoung Mu~Lee.
\newblock Deeply-recursive convolutional network for image super-resolution.
\newblock In {\em IEEE Conference on Computer Vision and Pattern Recognition},
  2016.

\bibitem[\protect\citeauthoryear{Kumar \bgroup \em et al.\egroup
  }{2009}]{kumar-iccv09-pubFig}
Neeraj Kumar, Alexander~C Berg, Peter~N Belhumeur, and Shree~K Nayar.
\newblock Attribute and simile classifiers for face verification.
\newblock In {\em IEEE International Conference on Computer Vision}, 2009.

\bibitem[\protect\citeauthoryear{Ledig \bgroup \em et al.\egroup
  }{2017}]{ledig-cvpr17-gan}
Christian Ledig, Lucas Theis, Ferenc Husz{\'a}r, Jose Caballero, Andrew
  Cunningham, Alejandro Acosta, Andrew Aitken, Alykhan Tejani, Johannes Totz,
  and Zehan Wang.
\newblock Photo-realistic single image super-resolution using a generative
  adversarial network.
\newblock In {\em IEEE Conference on Computer Vision and Pattern Recognition},
  2017.

\bibitem[\protect\citeauthoryear{Liu \bgroup \em et al.\egroup
  }{2005}]{liu-cvpr05-hf}
Wei Liu, Dahua Lin, and Xiaoou Tang.
\newblock Hallucinating faces: Tensorpatch super-resolution and coupled residue
  compensation.
\newblock In {\em IEEE Conference on Computer Vision and Pattern Recognition},
  2005.

\bibitem[\protect\citeauthoryear{Liu \bgroup \em et al.\egroup
  }{2007}]{liu-ijcv07-FH}
Ce~Liu, Heung-Yeung Shum, and William~T Freeman.
\newblock Face hallucination: Theory and practice.
\newblock {\em International Journal of Computer Vision}, 2007.

\bibitem[\protect\citeauthoryear{Liu \bgroup \em et al.\egroup
  }{2011}]{liu-pami11-siftflow}
Ce~Liu, Jenny Yuen, and Antonio Torralba.
\newblock Sift flow: Dense correspondence across scenes and its applications.
\newblock {\em IEEE Transactions on Pattern Analysis and Machine Intelligence},
  2011.

\bibitem[\protect\citeauthoryear{Lowe}{2004}]{lowe-ijcv04-sift}
David~G Lowe.
\newblock Distinctive image features from scale-invariant keypoints.
\newblock {\em International Journal of Computer Vision}, 2004.

\bibitem[\protect\citeauthoryear{Ma \bgroup \em et al.\egroup
  }{2010}]{xiang-pr10-FH}
Xiang Ma, Junping Zhang, and Chun Qi.
\newblock Hallucinating face by position-patch.
\newblock {\em Pattern Recognition}, 2010.

\bibitem[\protect\citeauthoryear{Petschnigg \bgroup \em et al.\egroup
  }{2004}]{Petschnigg-siggraph04-JBF}
Georg Petschnigg, Maneesh Agrawala, Hugues Hoppe, Richard Szeliski, Michael
  Cohen, and Kentaro Toyama.
\newblock Digital photography with flash and no-flash image pairs.
\newblock {\em ACM Transactions on Graphics (SIGGRAPH)}, 2004.

\bibitem[\protect\citeauthoryear{Song \bgroup \em et al.\egroup
  }{2014}]{song_eccv14_sketch}
Yibing Song, Linchao Bao, Qingxiong Yang, and Ming-Hsuan Yang.
\newblock Real-time exemplar-based face sketch synthesis.
\newblock In {\em European Conference on Computer Vision}, 2014.

\bibitem[\protect\citeauthoryear{Tappen and Liu}{2012}]{tappen-eccv12-bayesian}
Marshall Tappen and Ce~Liu.
\newblock A bayesian approach to alignment-based image hallucination.
\newblock In {\em European Conference on Computer Vision}, 2012.

\bibitem[\protect\citeauthoryear{Wang and Tang}{2005}]{wang-SMC05-hf}
Xiaogang Wang and Xiaoou Tang.
\newblock Hallucinating face by eigentransformation.
\newblock {\em IEEE Transactions on Systems, Man, and Cybernetics, Part C:
  Applications and Reviews}, 2005.

\bibitem[\protect\citeauthoryear{Wang \bgroup \em et al.\egroup
  }{2004}]{wang-tip04-SSIM}
Zhou Wang, Alan~Conrad Bovik, Hamid~Rahim Sheikh, and Eero~P Simoncelli.
\newblock Image quality assessment: from error visibility to structural
  similarity.
\newblock {\em IEEE Transactions on Image Processing}, 2004.

\bibitem[\protect\citeauthoryear{Wang \bgroup \em et al.\egroup
  }{2014}]{wang-ijcv14-survey}
Nannan Wang, Dacheng Tao, Xinbo Gao, Xuelong Li, and Jie Li.
\newblock A comprehensive survey to face hallucination.
\newblock {\em International Journal of Computer Vision}, 2014.

\bibitem[\protect\citeauthoryear{Wang \bgroup \em et al.\egroup
  }{2017}]{wang-tip17-bayesian}
Nannan Wang, Xinbo Gao, Leiyu Sun, and Jie Li.
\newblock Bayesian face sketch synthesis.
\newblock {\em IEEE Transactions on Image Processing}, 2017.

\bibitem[\protect\citeauthoryear{Yang \bgroup \em et al.\egroup
  }{2010}]{jianchao-tip10-scsr}
Jianchao Yang, John Wright, Thomas Huang, and Yi~Ma.
\newblock Image super-resolution via sparse representation.
\newblock {\em IEEE Transactions on Image Processing}, 2010.

\bibitem[\protect\citeauthoryear{Yang \bgroup \em et al.\egroup
  }{2013}]{Chih-cvpr13-FH}
Chih-Yuan Yang, Sifei Liu, and Ming-Hsuan Yang.
\newblock Structured face hallucination.
\newblock In {\em IEEE Conference on Computer Vision and Pattern Recognition},
  2013.

\bibitem[\protect\citeauthoryear{Yu and Porikli}{2016}]{yu-eccv16-ultra}
Xin Yu and Fatih Porikli.
\newblock Ultra-resolving face images by discriminative generative networks.
\newblock In {\em European Conference on Computer Vision}, 2016.

\bibitem[\protect\citeauthoryear{Zhou \bgroup \em et al.\egroup
  }{2015}]{zhou-aaai15-learning}
Erjin Zhou, Haoqiang Fan, Zhimin Cao, Yuning Jiang, and Qi~Yin.
\newblock Learning face hallucination in the wild.
\newblock In {\em AAAI Conference on Artificial Intelligence}, 2015.

\end{thebibliography}

\end{document}